\documentclass[sigconf]{acmart}
\copyrightyear{2025}
\acmYear{2025}
\setcopyright{cc}
\setcctype{by}
\acmConference[ICVGIP 2025]{Indian Conference on Computer Vision, Graphics, and Image Processing}{December 17--20, 2025}{Mandi, India}
\acmBooktitle{Indian Conference on Computer Vision, Graphics, and Image Processing (ICVGIP 2025), December 17--20, 2025, Mandi, India}
\acmPrice{}
\acmDOI{10.1145/3774521.3774555}
\acmISBN{979-8-4007-1930-1/2025/12}

\usepackage{booktabs} 
\usepackage{colortbl}
\usepackage{booktabs}
\usepackage{changepage}
\definecolor{mygray}{gray}{0.9}
\usepackage{tabularray}

\usepackage{framed}
\usepackage{xcolor}

\definecolor{shadecolor}{RGB}{50,50,50}          
\definecolor{prompttitlecolor}{RGB}{75,75,75}   
\definecolor{prompttextcolor}{RGB}{255,255,255} 

\newcounter{prompt}[section]
\renewcommand{\theprompt}{\thesection.\arabic{prompt}}

\newenvironment{prompt}[1][]{
    \refstepcounter{prompt}
    \begin{shaded*}
        \noindent
        {\colorbox{prompttitlecolor}{%
            \parbox{\dimexpr\linewidth-2\fboxsep}{%
                \color{white}\textbf{Prompt \theprompt: #1}%
            }%
        }}%
        \par\smallskip
        \color{prompttextcolor} 
}{
    \end{shaded*}
    \par\medskip
}

\begin{document}
\title{\underline{CSGaze}: \underline{C}ontext-aware \underline{S}ocial \underline{Gaze} Prediction}



\author{Surbhi Madan}
\email{surbhi.19csz0011@iitrpr.ac.in}
\orcid{0009-0000-3774-8117}
\affiliation{%
  \institution{Indian Institute of Technology Ropar}
 \city{Ropar}
 \country{India}
}

\author{Shreya Ghosh}
\email{shreya.ghosh@uq.edu.au}
\orcid{0000-0002-2639-8374}
\affiliation{%
  \institution{The University of Queensland}
 \city{Brisbane}
 \country{Australia}
}

\author{Ramanathan Subramanian}
\email{ram.subramanian@canberra.edu.au}
\orcid{0000-0001-9441-7074}
\affiliation{%
 \institution{University of Canberra}
 \city{Canberra}
 \country{Australia}
}

\author{Tom Gedeon}
\email{tom.gedeon@curtin.edu.au}
\orcid{0000-0001-8356-4909}
\affiliation{%
  \institution{Curtin University}
 \city{Perth}
 \country{Australia}
}

\author{Abhinav Dhall}
\email{abhinav.dhall@monash.edu}
\orcid{0000-0002-2230-1440}
\affiliation{%
  \institution{Monash University}
  \city{Melbourne}
  \country{Australia}
}

\renewcommand{\shortauthors}{Madan et al.}

\begin{abstract}
A person's gaze offers valuable insights into their focus of attention, level of social engagement, and confidence. In this work, we investigate how contextual cues combined with visual scene and facial information can be effectively utilized to predict and interpret social gaze patterns during conversational interactions. We introduce \textit{CSGaze}, a context-aware multimodal approach that leverages facial and scene information as complementary inputs to enhance social gaze pattern prediction from multi-person images. The model also incorporates a fine-grained attention mechanism centered on the principal speaker, which helps in better modeling social gaze dynamics. Experimental results show that CSGaze performs competitively with state-of-the-art methods on GP-Static, UCO-LAEO and AVA-LAEO. Our findings highlight the role of contextual cues in improving social gaze prediction. Additionally, we provide initial explainability through generated attention scores, offering insights into the model’s decision-making process. We also demonstrate our model's generalizability by testing our model on open set datasets that demonstrate its robustness across diverse scenarios. The code and data are available at \href{https://github.com/surbhimadan92/CS}{Link}.
\end{abstract}

%
%



\maketitle


\section{Introduction}
\begin{figure}[t]
     \centering
    \includegraphics[width = \linewidth]{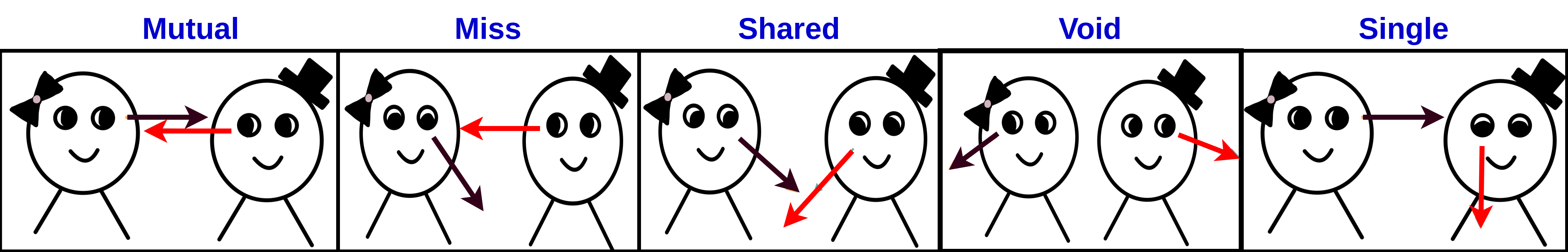}
    \caption{\small Illustration of static gaze patterns in dyadic communication. The five figures depict different gaze patterns, with the girl as the principal and the boy as the associate.} \label{fig: gaze_pattern}
    \vspace{-2mm}
\end{figure}

In social interactions, gaze is a crucial mechanism for expressing interest and conveying emotions. During communication, individuals continuously analyze, respond to, and observe their partner’s gaze patterns \cite{boucher2012reach}. Gaze also reflects intent—speakers often direct their gaze toward their speech target or the object they reference \cite{lee2024towards,ghosh2022av,ghosh2022mtgls}. Several studies have explored gaze target detection, which identifies the object a person is visually focusing on within an image \cite{so_120,so_121,so_122,so_115}. However, relatively few studies have explored computational approaches for analyzing social gaze \cite{chong2020detecting,gupta2024unified,malik2023relational}. Understanding social gaze patterns has broad applications, including developing AI systems with social intelligence \cite{gupta2024unified}, assessing developmental disorders such as Autism Spectrum Disorder (ASD) \cite{american2013diagnostic}, enhancing emotion understanding \cite{EM1,EM2}, and improving engagement prediction \cite{E1} and inferring human traits \cite{p1,P2} in interactive settings. Psychological studies show that the frequency and duration of different gaze patterns reveal whether a conversation flows smoothly or breaks down \cite{argyle1994gaze,kleinke1986gaze}. Therefore, automatically detecting and analyzing distinct gaze behaviors is essential for enabling machines to interpret communication between individuals.\\
Understanding human gaze communication in social interactions is essential for several reasons: (1) it sheds light on multi-agent gaze behaviours in real-world social settings, (2) it helps robots learn human gaze patterns for more intuitive and efficient interactions, (3) it enables realistic gaze behaviour simulations in Virtual Reality, (4) it contributes to studying mental states in social contexts, and (5) it supports autism diagnosis and assessment in children \cite{fan2019understanding}. Despite its significance, social gaze prediction remains challenging due to the absence of a standardized encoding framework for gaze behaviours. Most prior studies have focused on detecting isolated gaze patterns \cite{marin2014detecting,fan2018inferring,tafasca2023childplay}, limiting their ability to generalize and identify multiple simultaneous gaze interactions in images or videos. Fan et al. \cite{fan2019understanding} proposed six atomic-level gaze communication patterns based on cognitive psychology terminology. However, their framework lacks uniqueness, as three patterns (avert, refer, follow) represent sequential actions and can be decomposed into the remaining three (single, mutual, share), which describe static gaze states. Chang et al. \cite{gaze} attempted to address this issue by defining inseparable gaze patterns as stationary gaze states in dyadic interactions (Shown in Figure \ref{fig: gaze_pattern}), proposing that gaze behaviours in videos could be represented as sequences of frame-by-frame patterns.


Social gaze prediction has traditionally relied on specific visual cues such as head orientation, scene context, and the relative positions of individuals in an image. While these approaches provide valuable insights, they often lack a comprehensive understanding of the broader contextual elements that influence gaze behaviour. A possible solution to preserve contextual cues is manual annotation or pseudo-labeling using expert models; however, these methods are resource-intensive, require task-specific models for each cue, and may not always be feasible during inference. This highlights the need for more efficient and scalable solutions. Recently, the rise of Large Language Models (LLMs) \cite{Gaze4_33,Gaze4_34,Gaze4_12} has driven research into text-guided feature learning across various vision tasks, including crowd counting \cite{Gaze4_28}, point cloud analysis \cite{Gaze4_56}, and image generation \cite{Gaze4_39}. Despite these advancements, social gaze prediction has yet to fully harness the power of language-driven reasoning. We propose leveraging MLLMs to encode and preserve the contextual information of a scene, thereby enhancing social gaze understanding. By utilizing the zero-shot capabilities of MLLMs, we aim to incorporate linguistic context while minimizing computational overhead, paving the way for a more scalable and interpretable approach to social gaze prediction.\\




\begin{prompt}[MLLM (Kosmos-2) Interaction]
\textbf{System:} Initialize Kosmos-2.

\textbf{Human:} \{$<$prompt$>$ Describe how the persons are interacting in the scene."\}

\textbf{AI:} \{OUTPUT provides a textual description of the scene, detailing the objects, people, and their interactions. (See Figure~2)\}

\label{fig:prompt}
\end{prompt}

Building on previous findings, this work investigates the efficacy of MLLM in context-aware social gaze prediction. We introduce CS-Gaze, a carefully designed language-vision model aimed at enhancing the semantic space of social gaze estimation. Unlike traditional approaches that focus solely on refining image encoders, our method harnesses the guidance of textual signals to improve gaze prediction. Specifically, we leverage the zero-shot capabilities of MLLMs by defining a structured prompt: ``Describe how the persons are interacting in the scene." (as shown in Prompt 1.1). This prompt extracts meaningful scene descriptions, which are then integrated into the gaze prediction framework. To further refine the prediction process, we implement a fine-grained attention mechanism that prioritizes the principal person (whose gaze is being predicted) while also considering their interaction with an associate. Additionally, a cross-modal attention-based fusion approach is employed to effectively align textual and visual representations, allowing for a richer and more contextually aware learning process. By shifting the emphasis from image encoders to language-driven contextualization, our method enhances both interpretability and performance. We evaluate CS-Gaze on the GP-Static \cite{gaze} dataset and further test its generalizability on three additional datasets: UCO-LAEO \cite{uco}, AVA-LAEO \cite{ava} and VSGaze \cite{gupta2025mtgs}.

\begin{itemize}
    \item \textbf{Context-Guided Social Gaze Estimation:} \textbf{CSGaze} is the first to leverage MLLM-driven scene context for social gaze prediction, enabling a scalable and interpretable approach.
    
    \item \textbf{Fine-Grained Learning:} With attention mechanisms we focus more on the principal person, alongside cross-modal attention to effectively align textual and visual features.
    
    \item \textbf{State-of-the-Art Performance:} \textbf{CSGaze} outperforms existing methods on GP-Static dataset \cite{gaze} and generalizes well to UCO-LAEO \cite{uco}, AVA-LAEO \cite{ava}, and VSGaze \cite{gupta2025mtgs}, demonstrating the value of language-informed context in gaze understanding.
\end{itemize}

\section{Related Work}
This section examines recent advancements~\cite{ghosh2023automatic} in computer vision for identifying social gaze patterns and leveraging gaze behaviour to analyze complex social interactions.

\noindent \textbf{Single Gaze Pattern.}
Researchers have primarily focused on two major gaze patterns in the literature: Looking-At-Each-Other (LAEO) \cite{marin2014detecting}, also known as mutual gaze, and Shared Attention (SA) \cite{fan2018inferring}. Shared attention has been studied in two ways: (1) as a classification task—determining whether shared attention is present or not—and (2) as a localization task—identifying the object or region that two or more individuals are looking at. For LAEO detection, many approaches analyze cropped head images to extract gaze direction cues. These cues are then combined with either 2D spatial data or estimated 3D geometric features to determine the LAEO label \cite{doosti2021boosting,lin2017feature}. For shared attention, several methods have integrated gaze-following heatmap predictions from all individuals \cite{chong2020detecting,tu2022end}, leading to improved performance in shared attention prediction. However, existing approaches have certain limitations, such as the inability to distinguish between multiple shared attention instances occurring simultaneously and the computational complexity of processing each person pair independently.

\begin{figure*}[t]
   \centering
    \includegraphics[scale = 0.7]{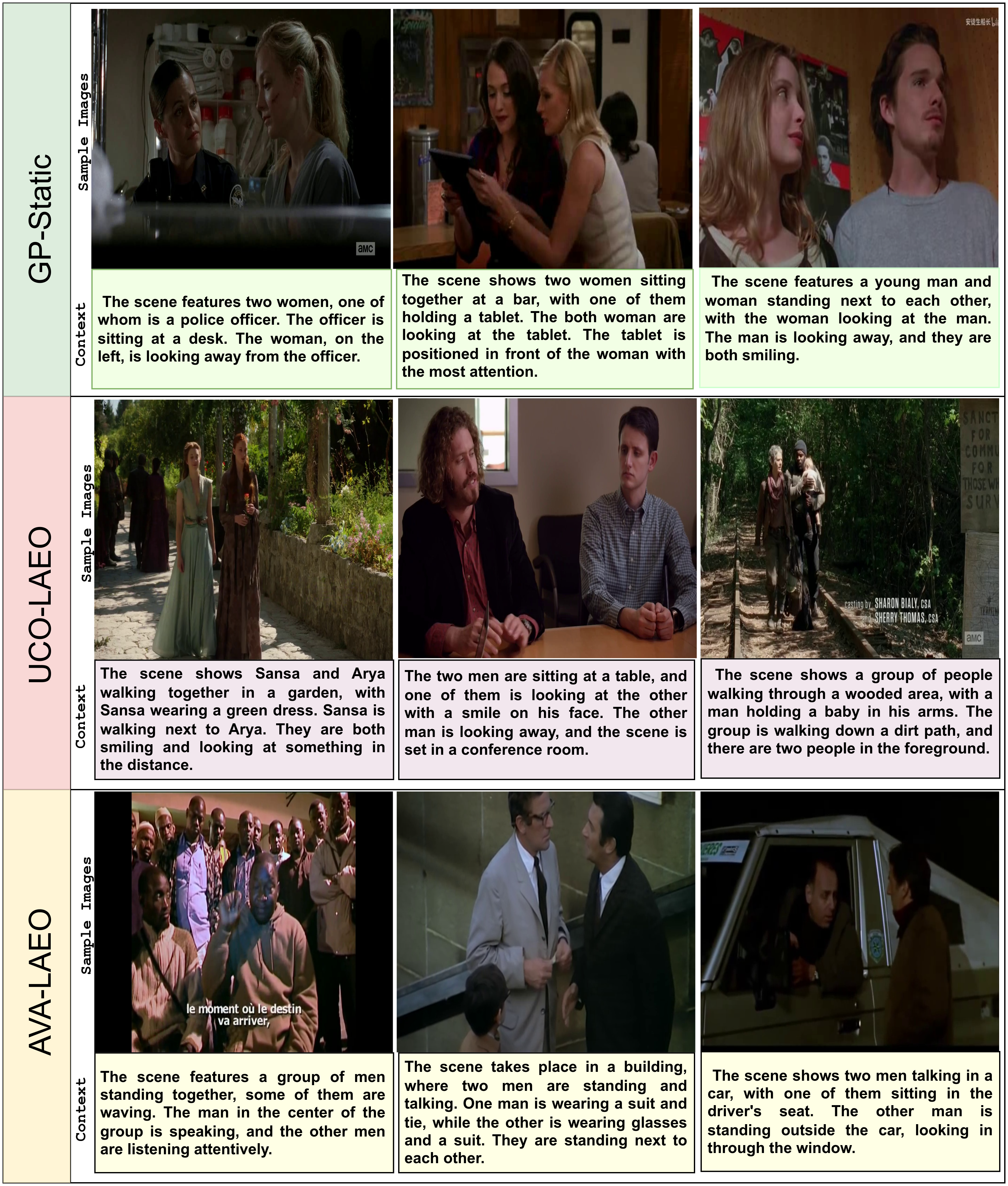}
    \caption{\small Sample frames with their contextual information from three datasets, Top: GP-Static, Middle: UCO-LAEO and End: AVA-LAEO.} 
    \label{fig: mllm}
    \vspace{-2mm}
\end{figure*}

\begin{figure*}[t]
    \centering
    \includegraphics[scale = 0.8]{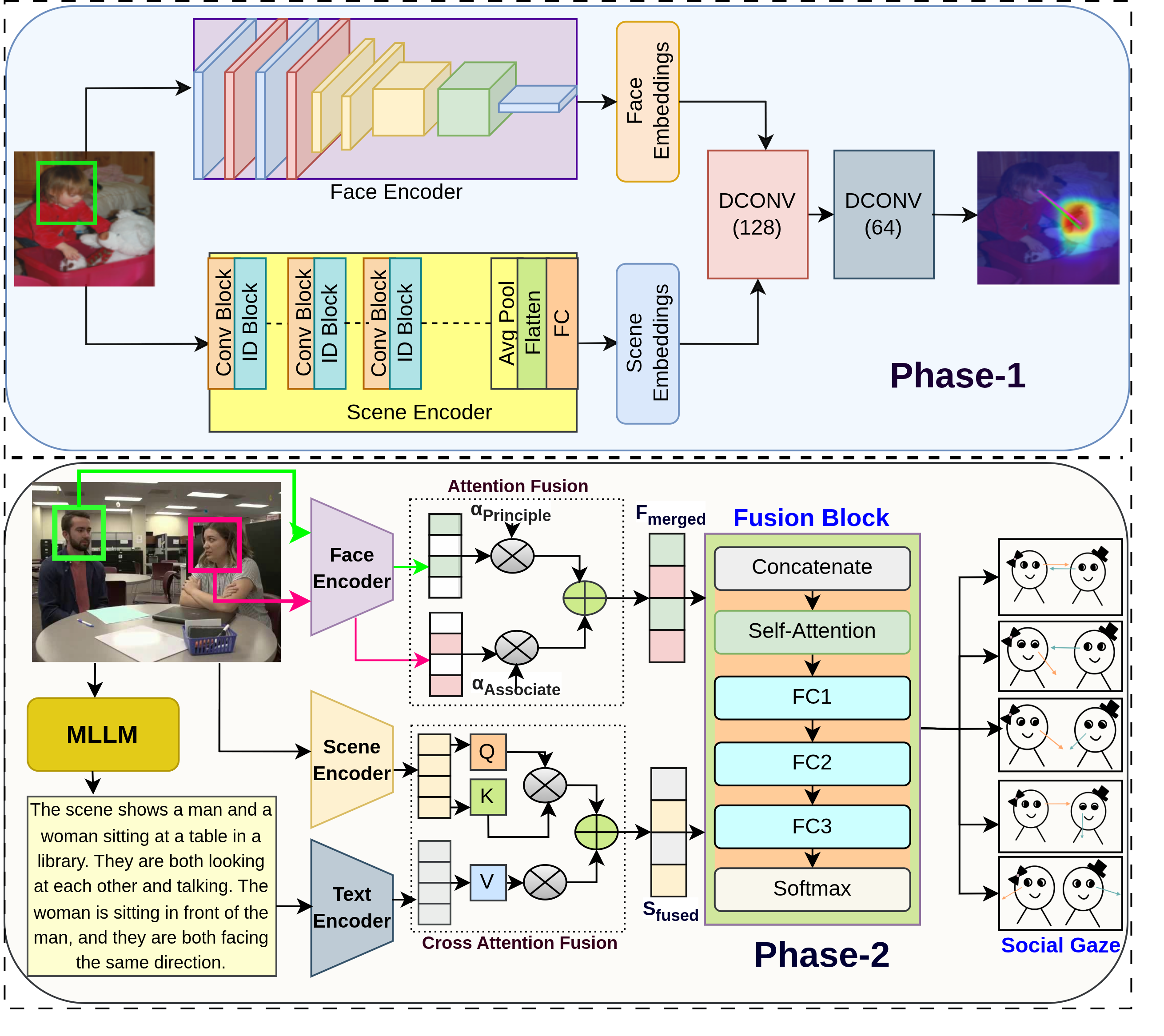}
    \vspace{-3mm}
    \caption{\small \textbf{CS-Gaze Framework.} Phase 1: We pre-train the scene and face encoders using the GazeFollow dataset \cite{nips15_recasens}, which contains images annotated with gaze locations, providing supervisory signals for gaze estimation (See Heatmap, headlocation and gaze vector). Stage 2: Given an image with two interacting individuals: \textit{Principal} (Green bounding box) and \textit{Associate} (Pink bounding box), their cropped facial regions are processed through the pretrained face encoder to extract face-specific embeddings. To model the interaction between the two individuals, we employ an attention-based fusion mechanism, where the embeddings of both individuals are combined using learnable weighting parameters that determine the relative importance of each person’s facial features in the final representation.} 
    \label{fig: overall}
    \vspace{-3mm}
\end{figure*}

\noindent \textbf{Multiple Gaze Communication Patterns.}
Multi-gaze analysis remains an underexplored area, with single gaze pattern analysis facing notable limitations, such as challenges in generalizing across tasks and the complexity of processing person pairs independently. These approaches often fail to capture the full spectrum of human gaze behaviour. Despite the significance of multi-gaze prediction, only a few studies have addressed this topic \cite{fan2019understanding,gaze,gupta2025mtgs,gupta2024unified}. Fan et al. \cite{fan2019understanding} proposed a spatio-temporal graph network to hierarchically reason about gaze communication at both the atomic level (e.g., shared attention, mutual gaze) and the event level (e.g., gaze aversion, joint attention). However, their approach assumes atomic-level gaze behaviours are mutually exclusive, which is an oversimplification. For instance, person A can simultaneously maintain mutual gaze with person B while sharing attention with person C towards B.\\ Chang et al. \cite{gaze} redefined five static gaze patterns encompassing all possible gaze states in dyadic interactions. They introduced a network for recognizing these mutually exclusive gaze patterns from a single image and proposed the GP-Static dataset to support further research. Meanwhile, Gupta et al. \cite{gupta2025mtgs} developed VSGaze, a dataset that provides annotations for both social gaze and gaze following, facilitating the training of more generalizable models. The social gaze labels in VSGaze include Looking-at-Head (LAH), LAEO, and SA, alongside precise gaze point annotations. Gupta et al. \cite{gupta2024unified} introduce a unified model for gaze following and social gaze prediction by using Graph Attention Network to explicitly model people interactions, before being passed to the pairwaise task heads. To the best of our knowledge, these are the only studies focused on multi-gaze analysis. Notably, while GP-Static is publicly available, VSGaze remains inaccessible at the time of writing.

\noindent \textbf{Context-Aware Social Gaze Prediction.}
Visual gaze estimation has gained significant research interest over the past decade due to its diverse applications. While existing methods have improved prediction accuracy, they primarily rely on single-image cues to infer gaze direction, often neglecting the potential of text-based guidance, which has become increasingly relevant in vision-related tasks. To date, only two studies have explored this integration. Ruan et al. \cite{wang2023gazeclip} proposed GazeCLIP, a text-guided gaze estimation approach leveraging the strong generalization capability of the CLIP model across vision tasks. This work was the first to distill knowledge from a large-scale vision-language model to enhance gaze estimation learning. Meanwhile, Gupta \emph{et al. }\cite{gupta2024exploring} examined the zero-shot capabilities of vision-language models (VLMs) for gaze following. Their study systematically evaluates different VLMs, prompting strategies, and in-context learning (ICL) techniques to analyze zero-shot cue recognition. They further extract contextual cues for gaze following and assess their impact when incorporated into a state-of-the-art gaze prediction model.

\section{Methodology}

\subsection{Problem Formulation}

Given an image \( I \) containing two individuals, \( P_1 \) and \( P_2 \), where \( P_1 \) is designated as the \textbf{Principal} (whose gaze is to be predicted) and \( P_2 \) is the \textbf{Associate} (engaged in conversation with \( P_1 \) in a dyadic setting), the goal is to estimate the probability distribution over \( N \) possible \textbf{social gaze classes} for \(P_1 \) . This is formulated as a \textbf{multiclass classification problem}, where the model predicts:  

\[
\mathbf{y} = [y_1, y_2, \dots, y_N]
\]

such that:

\[
y_i = P(C_i \mid I, P_1, P_2), \quad \forall i \in \{1, 2, \dots, N\}
\]

where \( C_i \) represents the \( i^{th} \) social gaze class. The predicted output \( \mathbf{y} \) provides the likelihood of each class occurring in the given conversational scenario.





\subsection{Features Extraction}
\subsubsection{Face Features}
To analyze the social gaze patterns of the principal participant in a conversation, it is essential to encode the facial and head features of both participants. To achieve this, we first crop the faces of both individuals and pass them through a dedicated face encoder to extract their respective head-related features. The model employs a convolutional pathway to implicitly capture key information regarding head poses and gaze angles, ensuring a robust feature representation for gaze prediction.

\subsubsection{Scene Features}
Building upon prior studies \cite{chong2020detecting,gaze,so_115}, which highlight the role of saliency information in gaze estimation, we introduce a scene branch to incorporate visual context from the surrounding environment. This branch processes the entire image as input and extracts scene features that complement gaze understanding, capturing crucial background elements that influence social gaze behaviour.

\subsubsection{Contextual Features}
Given the remarkable advancements of MLLMs across various domains, we explore their potential to encode contextual cues within a scene. By providing an input image along with a structured prompt, the MLLM generates a descriptive textual summary of the scene. Some of the samples MLLM generated contextual information is shown in Figure \ref{fig: mllm}. To effectively integrate this information, we encode the generated textual descriptions into vector representations using a text encoder. This additional modality, combined with face and scene features, enhances the model’s ability to interpret social interactions in conversational settings. 




\subsection{Proposed Method}
To effectively predict and analyze social gaze patterns in conversational settings, we introduce CSGaze, a context-aware social gaze prediction framework. Our approach leverages multi-modal information, integrating facial features, visual scene context, and overall conversational cues to make robust gaze predictions (See Figure \ref{fig: overall}). 

\noindent In the \textbf{first phase}, we pre-train the scene and face encoders using the GazeFollow dataset \cite{nips15_recasens}, which contains images annotated with gaze locations, providing supervisory signals for gaze estimation. This pretraining step enables the encoders to learn spatially and contextually aware feature representations, capturing both the individual’s facial and head features and scene-dependent gaze cues. The model is trained to predict a $64 \times 64$ heatmap representing the probability distribution of the gaze target within the image. It ensures that the encoders develop a foundational understanding of gaze behavior before being adapted for the conversational setting. 

\noindent Once the encoders are pretrained, the \textbf{second phase} focuses on context-aware social gaze classification. Given an image with two interacting individuals: \textbf{Principal} and \textbf{Associate}, their cropped facial regions are processed through the pretrained face encoder to extract face-specific embeddings. To model the interaction between the two individuals, we use an attention-based fusion mechanism that combines their embeddings using learnable weights, $\alpha_{\text{Principal}}$ and $\alpha_{\text{Associate}}$, which determine the relative importance of each person's facial features in the final representation. This final representation is marked as $F_{\text{merged}}$. Next, contextual cues are generated using a MLLM to preserve the overall context of the input image. The visual scene is processed using the trained scene encoder from Phase 1, while contextual feature embeddings are extracted using a text encoder. These embeddings, representing visual and contextual aspects, are fused using a cross-attention mechanism termed $S_{\text{fused}}$, to ensure the model attends to the most relevant information in the scene for gaze prediction. The fused facial embeddings ($F_{\text{merged}}$) and scene-context embeddings ($S_{\text{fused}}$) are further combined using self-attention, allowing the model to capture interactions and dependencies between the modalities. This final merged representation is then passed through fully connected layers, followed by an activation layer, to predict the static gaze class of the given input image. The overall framework ensures that CSGaze effectively captures both individual and contextual factors influencing social gaze behavior in conversational settings. 


\section{Experimental Setup and Results}

\subsection{Datasets}
We conducted our experiments on multiple datasets as: \\
\noindent \textbf{GP-Static Dataset:} The GP-Static dataset \cite{gaze} contains 84,682 frames with 169,364 frame-level static gaze annotations, where each individual is annotated per frame. Around 20\% (14,952 frames) is reserved for testing, with no overlap in TV shows or movies between training and test sets to ensure evaluation on unseen data. It defines five social gaze classes: Share, Mutual, Single, Miss, and Void (see Figure \ref{fig: gaze_pattern}).


\begin{figure}[!h]
    \centering
    \includegraphics[scale = 0.3]{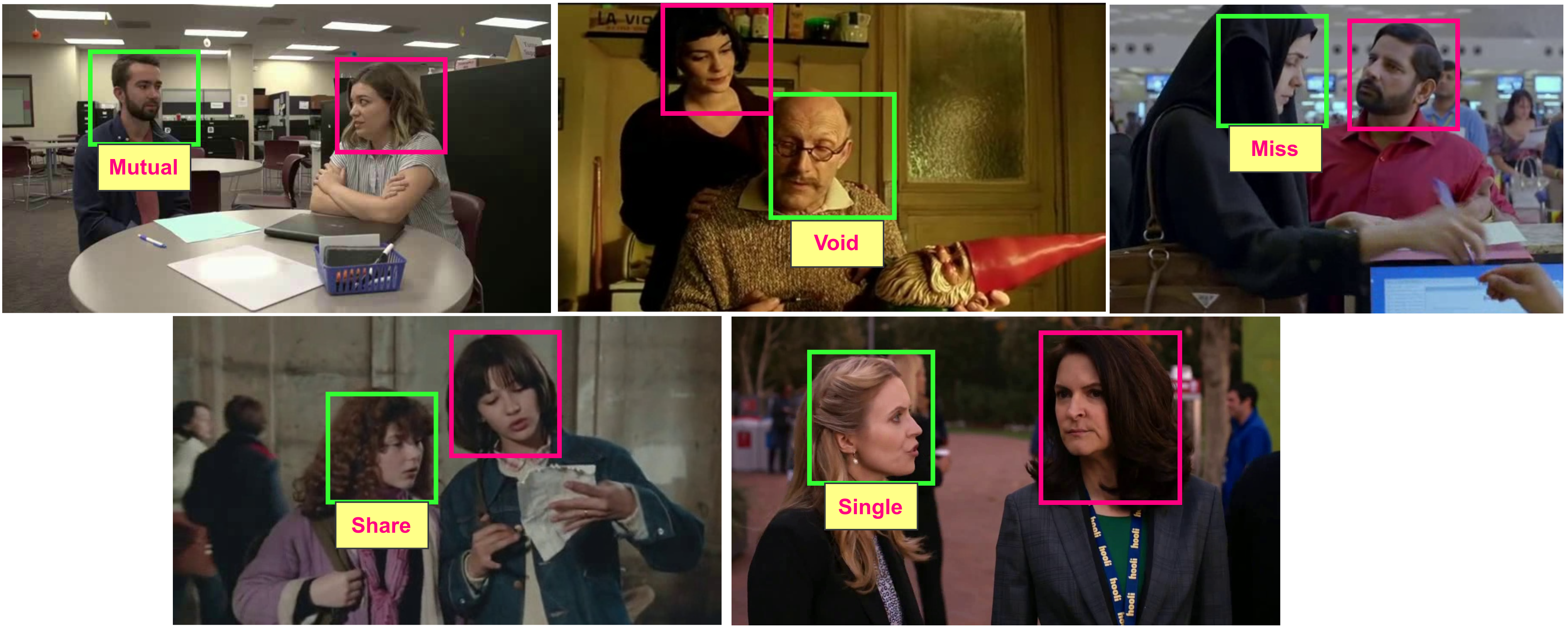}
    \caption{\small Sample images from the GP-Static dataset for each corresponding class label. } 
    \label{fig: dataset_frames}
    \vspace{-2mm}
\end{figure}

\noindent \textbf{VSGaze Dataset:} The VSGaze dataset includes annotations for three social gaze classes: LAEO, LAH (Looking At Head) and SA. As VSGaze is not publicly available, we reconstructed it following \cite{gupta2025mtgs}, using over 800k frame samples sourced from VideoCoAtt \cite{fan2018inferring}, UCO-LAEO \cite{uco}, ChildPlay \cite{tafasca2023childplay}, and VideoTargetAttention (VAT) \cite{chong2020detecting}. As each dataset annotates only a subset of individuals, we detected missing heads using a pre-trained YOLOv5 detector and predicted gaze points with Gaze-LLE \cite{ryan2025gaze}. LAH pairs were identified when a person’s gaze intersects another’s head bounding box, LAEO pairs via mutual gaze detection, and SA pairs when both individuals’ gaze targets fall within the same head box. \\

\noindent \textbf{UCO-LAEO and AVA-LAEO:} To assess our model on single-gaze tasks, we use two datasets for the LAEO label: UCO-LAEO and AVA-LAEO. UCO-LAEO contains 22,398 images from four TV shows, with 6,114 positive LAEO pairs among 36,740 possible head pairs, each annotated with bounding boxes and LAEO labels. AVA-LAEO, derived from AVA v2.2, includes ~19k LAEO and ~118k not-LAEO pairs in the training set, and ~5.8k LAEO and ~28k not-LAEO pairs in the validation set. Each human pair is frame-level annotated as LAEO, not-LAEO, or ambiguous, offering a robust benchmark for LAEO detection.

\begin{table}[t]
\centering
\renewcommand{\arraystretch}{1.5}
\fontsize{7}{7}\selectfont
\caption{\small Performance comparison of SOTA methods using classwise F1 score on \textit{GP-Static} test set.}
\label{tab:classwise}   
\resizebox{8.8cm}{!}{
\begin{tabular}{l||ccccc|cc}
\hline
\textbf{Methods}   & \multicolumn{5}{c|}{\textbf{Social Gaze Classes (F1 Score $\uparrow$)}}  &  \multicolumn{2}{c}{\textbf{5-Class Avg.}}\\ \cline{2-6}
\textbf{(GP-Static)}   & \textbf{Mutual} & \textbf{Single} & \textbf{Miss} & \textbf{Void} & \textbf{Share} & \textbf{F1 $\uparrow$} & \textbf{Acc.$\uparrow$}  \\
\hline \hline

GF Fixed \cite{chong2020detecting}    & 0.46            & 0.31            & 0.31          & 0.36          & 0.18            & 0.32                & 0.35              \\
GF Modified \cite{chong2020detecting} & 0.61            & 0.26            & 0.29          & 0.42          & 0.34            & 0.38                & 0.43              \\
Baseline \cite{gaze}    & 0.79            & 0.59            & 0.59          & 0.6           & \textbf{0.73 }          & 0.66                & 0.67              \\
CSGaze (Ours)        & \textbf{0.83  }          & \textbf{0.64 }           & \textbf{0.63 }         & \textbf{0.61 }         & 0.69      & \textbf{0.68 }               & \textbf{0.69 }     \\           
\hline
\end{tabular}}
\end{table}

\subsection{Prediction Settings} We explore both multiclass and binary classification for social gaze prediction. Results are presented in Tables~\ref{tab:classwise}, \ref{tab:sota} and \ref{tab:vsgaze}. During training, our model is fine-tuned using 10\% of the training data for validation and evaluated on the test set. For consistency with prior work, we report the following metrics: F1-score and average accuracy for GP-Static; F1-score and average precision for VSGaze; and average precision across 100 runs for UCO-LAEO and AVA-LAEO.\\

\noindent \textbf{Performance Metrics.} We evaluate the performance of the social gaze pattern classification task using F1-score and Average Accuracy, defined as the proportion of correctly classified gaze pattern instances. In addition, we report the Average Precision, which reflects the mean precision across varying recall levels and is commonly represented as the area under the precision-recall curve.

\subsection{Implementation Details}
We have conducted our experiments on an NVIDIA A100-40GB GPU using the TensorFlow framework. Given the coordinate values of the face, we have used the OpenCV Python library to crop facial regions.  The input images, including cropped face images of both individuals, are resized to 224×224. For feature extraction, we employ ResNet34 \cite{resnet34} as the face encoder and ResNet-50 as the scene encoder. KOSMOS-2 \cite{kosmos} served as our MLLM, while RoBERTa \cite{roberta} is used as the text encoder to provide textual embeddings. For the LAEO datasets, since both participants contribute equally to the interaction, equal attention weights (0.5 each) are assigned to the principal ($\alpha_{\text{Principal}}$) and associate ($\alpha_{\text{Associate}}$) participants. The training process is divided into two phases: in Phase-1, the model is pretrained for 10 epochs. During Phase-2 training, we have applied categorical cross-entropy loss for multiclass classification and binary cross-entropy loss for binary classification. The network is optimized using the Adam optimizer with a learning rate of 0.001, and the batch size was set to 128. The model is trained for 200 epochs, with early stopping applied, using a patience value of 5.

\begin{figure*}[t]
\centering
\includegraphics[scale = 0.65]{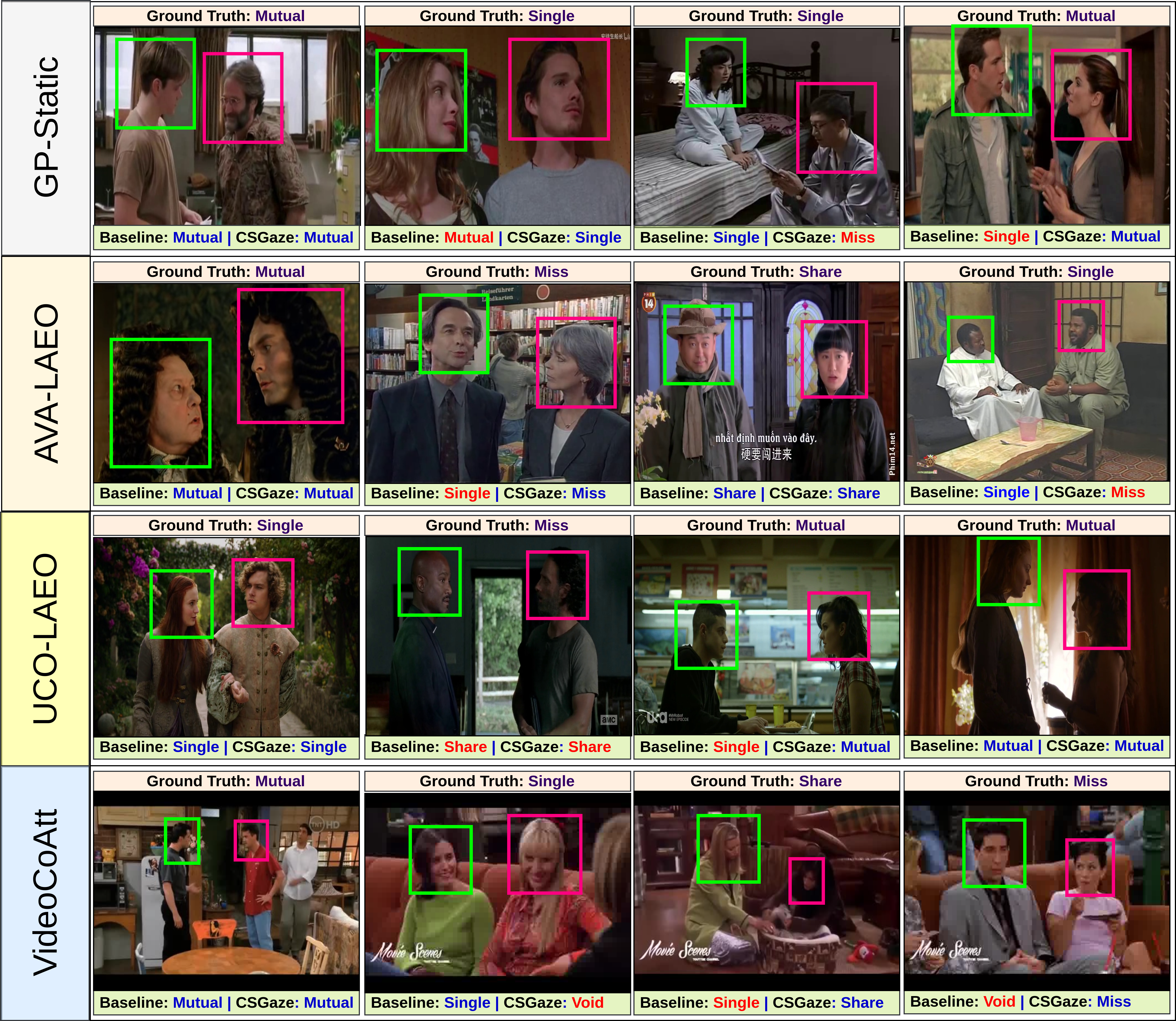}
\caption{\small Qualitative Analysis. Comparison of our proposed approach, CSGaze, with the baseline method \cite{gaze} across four diverse datasets: GP-Static, AVA-LAEO, UCO-LAEO, and VideoCoAtt. In each example, green bounding boxes indicate the principal face, while pink boxes denote the associate. Predictions are color-coded: blue for correct and red for incorrect outputs.}
\label{qualitative}
\end{figure*}

\subsection{Results and Discussion}

\subsubsection{Overall and Classwise Prediction Results} Table~\ref{tab:classwise} presents the overall classification results on the GP-Static test set. Our proposed approach, CSGaze, outperforms other methods, including the baseline, achieving a 2\% improvement in both average accuracy and average F1 scores. Although the gain appears incremental, it highlights the inherent challenges in recognizing social gaze patterns.


We present the classwise results on the test set of the GP-Static dataset in Table~\ref{tab:classwise}. Our proposed approach, CSGaze, outperforms the baseline across all classes except Share, achieving relative improvements of 5\%, 8.4\%, 6.8\%, and 2\% for the Mutual, Single, Miss, and Void classes, respectively. The lower performance in the Share class may be attributed to its reliance on scene-object interactions, where individuals attend to the same object or space, making it more complex compared to other classes that primarily depend on directional gaze focus. Despite this, our approach demonstrates an overall performance improvement.

\begin{table}[t]
\centering
\fontsize{7}{7}\selectfont
\renewcommand{\arraystretch}{1.5}
\caption{\small Performance comparison of SOTA methods using Average Precision (AP) metric on the test sets of the \textit{UCO-LAEO} and \textit{AVA-LAEO} datasets. (LAEO: Looking at Each Other). Please note that FG'24~\cite{gupta2024exploring} codebase is not publicly available.}
\label{tab:sota}   
 \resizebox{7cm}{!}{
 \begin{tabular}{l||cc}
\hline
\textbf{Methods} & \multicolumn{2}{c}{\textbf{LAEO Datasets (AP $\uparrow$)}} \\ \cline{2-3}
 & \textbf{UCO-LAEO} & \textbf{AVA-LAEO} \\
\hline \hline
LAEO-Net \cite{marin2011here}  & 79.5              & 50.6              \\
AAAI’21 \cite{doosti2021boosting}   & 65.1              & 72.2              \\
GP-Static \cite{gaze} & 80.3              & 82.5              \\
FG'24 \cite{gupta2024unified}      & 95.7              & -                 \\
MGTR \cite{mgtr} & 64.8 & 66.2\\
MGTR + GTD-LLM \cite{gtd} & 68.3&69.5 \\
LAEO-Net + GTD-LLM \cite{gtd} & 83.0 & 60.4 \\
CSGaze (Ours)           & 88.7              & \textbf{88.2}      \\         
\hline
\end{tabular}}
\end{table}

\subsubsection{Comparison with State-of-the-Art Approaches for the LAEO Class} To evaluate the effectiveness of our proposed approach, CSGaze, across diverse scenarios, we have tested it on the LAEO class using two benchmark datasets: UCO-LAEO and AVA-LAEO. Table \ref{tab:sota} compares CSGaze with state-of-the-art methods. CSGaze outperforms all baselines on the AVA-LAEO dataset, demonstrating the effectiveness of contextual information for social gaze prediction. On UCO-LAEO, it performs slightly below the recent method by Gupta et al. \cite{gupta2024unified}, likely due to their explicit modeling of gaze interactions and use of multimodal cues like speaking status. Notably, the CSGaze model is lightweight, with approximately 54M parameters, compared to \cite{gupta2024unified} (>100M). Compared to the baseline \cite{gaze} (~51M), CSGaze requires slightly more parameters due to the inclusion of cross-attention and contextual modules. Despite this, CSGaze surpasses other methods in multiple aspects, confirming its robustness and generalizability. We believe it is also suitable for real-time deployment.

\begin{table}[t]
\centering
\fontsize{7}{7}\selectfont
\renewcommand{\arraystretch}{1.5}
\caption{\small Performance comparison of SOTA methods using F1 score on \textit{VSGaze dataset}. (LAH: Looking at Head, SA: Shared Attention, LAEO: Looking at Each Other). Please note that the labels of VSGaze is not publicly available.}
\label{tab:vsgaze}   
\resizebox{7cm}{!}{
\begin{tabular}{l||ccc}
\hline
\textbf{Methods} & \textbf{LAH} & \textbf{LAEO} & \textbf{SA} \\ \cline{2-4}
\textbf{(VSGaze)} & \textbf{F1\textsubscript{LAH}$\uparrow$} & \textbf{F1\textsubscript{LAEO}$\uparrow$} & \textbf{AP\textsubscript{SA}$\uparrow$} \\
\hline  \hline
Chong\textsubscript{S} \cite{chong2020detecting}  & 0.78              & 0.56      & 0.29          \\
Chong\textsubscript{T} \cite{chong2020detecting}  & 0.76              & 0.53      & 0.33          \\
Gupta \cite{gupta2022modular}  & 0.78              & 0.59      & 0.36          \\
Baseline (MTGS) \cite{gupta2025mtgs} &0.81 &0.60 &0.58 \\
CSGaze (Ours)& \textbf{0.85} &\textbf{0.68} & \textbf{0.79}\\      
\hline
\end{tabular}}
\vspace{-5mm}
\end{table}

\begin{table*}[t]
\centering
\caption{\small Comparison of classification performance (F1 Score $\uparrow$) across different class combinations (2-class: Miss+Single, Mutual+Share; 3-class: Mutual+Share+Void, Miss+Void+Single; 4-Class: Miss+Mutual+Void+Single, Mutual+Share+Void+Single) on the \textit{GP-Static} dataset.}
\renewcommand{\arraystretch}{1.5}
\resizebox{17.5cm}{!}{
\begin{tabular}
{l||c c||c c||c c c||c c c||c c c c||c c c c} 
\hline
         & \multicolumn{4}{c||}{\textbf{2-Class}} & \multicolumn{6}{c||}{\textbf{3-Class}}  & \multicolumn{8}{c}{\textbf{4-Class}}  \\ 
\hline \hline
\textbf{Classes}  & \textbf{Miss} & \textbf{Single}   & \textbf{Mutual} & \textbf{Share}               & \textbf{Mutual} & \textbf{Share} & \textbf{Void  }      & \textbf{Miss} & \textbf{Void} & \textbf{Single}         & \textbf{Miss} & \textbf{Mutual} & \textbf{Void} & \textbf{Single} & \textbf{Miss} & \textbf{Share} & \textbf{Void} & \textbf{Single}  \\ 
\hline
\textbf{F1-Score} & 0.84 & 0.84                  & 0.97   & 0.95                & 0.94   & 0.78  & 0.8         & 0.73 & 0.72 & 0.73           & 0.64 & 0.79   & 0.69 & 0.65   & 0.71 & 0.68  & 0.63 & 0.71    \\
\hline
\end{tabular}
\label{tab:gp_static_results}
}
\end{table*}

\subsubsection{Comparison with State-of-the-Art Approaches for the VSGaze Dataset} Table~\ref{tab:vsgaze} compares our method with state-of-the-art approaches on the VSGaze dataset. Unlike prior models that rely on temporal dynamics, our method uses only static frames, simplifying inference and improving robustness. Despite not using temporal cues, CSGaze outperforms all baselines, including video-based models, demonstrating the effectiveness of our multi-modal fusion strategy. These results highlight CSGaze’s suitability for scenarios lacking video or consistent temporal information.

\subsubsection{Qualitative Analysis} Figure \ref{qualitative} presents a performance comparison between our proposed approach CSGaze and the baseline method \cite{gaze} across four diverse datasets: GP-Static, AVA-LAEO, UCO-LAEO, and VideoCoAtt. The samples shown are representative examples chosen to illustrate the model’s comparative performance. The results demonstrate that our approach effectively captures social gaze patterns across various scenarios.

\subsubsection{Ablation Study}

\textbf{Comparative Analysis of Prediction Performance Using Single and Combined Feature Modalities:} Table~\ref{tab:ablation} represents the results for different combinations of input features used in the social gaze prediction task. Individually, the scene (S) and context (C) features yield relatively modest F1 scores of 0.29 and 0.31, respectively, indicating limited discriminative power when used alone. Their combination (Scene+Context) shows a moderate improvement to 0.39, suggesting some complementary information between the two. However, a significant performance boost is observed when facial features (F\textsubscript{merged}) are introduced, achieving an F1 score of 0.60. Combining facial features with scene or contextual information further improves the performance to 0.62 and 0.64, respectively, demonstrating that facial features are the most informative modality, and both scene and contextual cues offer additional complementary benefits. The highest performance is achieved with the all three features, reaching an F1 score of 0.68, confirming that integrating all three modalities leads to the most comprehensive and effective representation for social gaze prediction. A similar trend is observed for accuracy, further reinforcing the value of multi-modal fusion in improving prediction quality.

\begin{table}[!h]
\centering
\fontsize{7}{7}\selectfont
\renewcommand{\arraystretch}{1.2}
\caption{\small Performance comparison of the proposed method (CSGaze) using F1 score and Average Accuracy on the \textit{GP-Static dataset}. \textbf{Scene} refers to embeddings from the Scene Encoder, \textbf{Context} is the textual description generated by the MLLM and encoded using Text Encoder, and \textbf{Face} represents the attention-fused facial features of both the Principal and Associate (F\textsubscript{merged}). \textbf{Scene+Context} denotes the fused scene and context features (S\textsubscript{fused}), while \textbf{Face+Scene+Context} corresponds to the full CSGaze framework.}

\label{tab:ablation}   
\resizebox{8cm}{!}
{

\begin{tabular}{l||cc}
\hline
\textbf{Modality} & \textbf{F1 $\uparrow$} & \textbf{Accuracy $\uparrow$}  \\ \cline{1-3}
Scene  & 0.29              & 0.34          \\
Context  & 0.31              & 0.38         \\
Scene $+$ Context & 0.39              & 0.43          \\
Face & 0.60             & 0.62          \\
Face $+$ Scene  & 0.62              & 0.63          \\
Face $+$ Context & 0.64              & 0.66          \\
Face $+$ Scene $+$  Context & 0.68              & 0.69          \\
    
\hline
\end{tabular}
}
\end{table}

\noindent \textbf{Comparison of results across various class combinations and reductions.} To better understand class-specific performance, we evaluated various class combinations on the GP-Static and VSGaze datasets. Results in Tables~\ref{tab:gp_static_results} and \ref{tab:vsgaze_results} yield the following insights. From Table~\ref{tab:gp_static_results}, performance declines as the number of classes increases, indicating greater difficulty in distinguishing between them. Notably, Mutual is consistently the easiest to recognize. In Table~\ref{tab:vsgaze_results}, Shared Attention (SA) emerges as the most easily classified class, likely due to the presence of group settings and contextual cues such as objects and scene interactions. In contrast, LAEO shows inconsistent performance, likely due to fewer samples. The most challenging distinction is between LAEO and LAH, with the lowest F1-score (0.38). Overall, classification accuracy depends heavily on class pairing—while some (e.g., SA vs. LAEO) are easier to distinguish, others (e.g., LAH vs. LAEO) cause more confusion. These findings emphasize that although more classes increase model complexity, multiclass classification offers better generalizability for real-world scenarios than single-class models.

\begin{table}[t]
\centering
\caption{\small Comparison of classification performance (F1 Score $\uparrow$) across different social gaze class combinations (i.e. LAH+LAEO, LAH+SA, LAEO+SA) on the VSGaze dataset.}
\resizebox{8cm}{!}{
\begin{tabular}{l||c c||c c||c c} 
\hline
\textbf{Social Gaze}         & \multicolumn{6}{c}{\textbf{2-Class}} \\ \cline{2-7} 

\textbf{Classes}  & \textbf{LAH}  & \textbf{LAEO}                  & \textbf{LAH}  & \textbf{SA}                    & \textbf{LAEO} & \textbf{SA}    \\ \hline \hline 
\textbf{F1-Score} & 0.87 & 0.38                  & 0.45 & 0.8                   & 0.61 & 0.94                   \\
\hline
\end{tabular}}
\label{tab:vsgaze_results}
\end{table}


\subsection{Initial Explainability and Discussion} For the initial explanations, we examine attention scores for two representations: Scene fused ($\mathbf{S}_{\text{fused}}$) as shown in Table \ref{fig: att_sco}, which integrates scene and text features to provide a holistic view of the image, and Face merged ($\mathbf{F}_{\text{merged}}$), which captures fine-grained fused representations from both participants on the GP-Static dataset.
\begin{table}[t]
\centering
\fontsize{7}{7}\selectfont
\renewcommand{\arraystretch}{1.5}
\caption{\small Mean modality-wise attention scores per class (5-classes for Social Gaze) on the GP-Static dataset, indicating the relative contribution of each modality during training.}
\label{fig: att_sco}
\resizebox{8cm}{!}{
\begin{tabular}{l||ccccc}
\hline
 & \textbf{Single} & \textbf{Void} & \textbf{Miss} & \textbf{Mutual} & \textbf{Share}  \\ 
\hline \hline
\textbf{$\mathbf{S}_{\text{fused}}$ }  & 0.4    & \textbf{0.65} & 0.45 & 0.28   & \textbf{0.59}   \\ 
\hline
\textbf{$\mathbf{F}_{\text{merged}}$ }   &\textbf{ 0.6  }  & 0.35 & \textbf{0.55} &\textbf{ 0.72}   & 0.41   \\
\hline
\end{tabular}}
\vspace{-5mm}
\end{table}

Our observations indicate that facial features, which incorporate head and eye information, are more predictive for the \emph{mutual}, \emph{single}, and \emph{miss} gaze classes. This is likely because the gaze target in these cases is always another person. In contrast, scene features contribute more significantly to the \emph{share} and \emph{void} classes. \emph{Share gaze} poses a greater challenge, as the gaze targets of both participants can be either another person or an object. This requires scene-object interaction, making the features of the scene more relevant - aligning with the findings of \cite{gupta2025mtgs}. Thus, incorporating additional contextual information enhances the accuracy of social gaze prediction.

\section{Conclusion}
In this paper, we present CSGaze, a novel approach for predicting social gaze patterns from images. Our method integrates contextual cues and a fine-grained attention mechanism focused on the principal speaker to enhance the accuracy of social gaze prediction. Through extensive experiments, we observe that while learning single-class gaze patterns is relatively easy, it is less applicable to real-world scenarios. In contrast, multi-class gaze pattern learning is more challenging but better reflects the complexities of real-world interactions. To support interpretability, we provide initial explainability through generated attention scores that offer insights into the model’s focus during prediction. For future work, we aim to extend this framework by incorporating temporal dynamics and enabling multi-person social gaze prediction to better capture the richness of real-world social interactions.

\bibliographystyle{ACM-Reference-Format}
\bibliography{ICVGIP-Latex-Template}





\end{document}